\pdfoutput=1

\documentclass[11pt]{article}

\usepackage{EMNLP2023}

\usepackage{times}
\usepackage{latexsym}

\usepackage[T1]{fontenc}

\usepackage[utf8]{inputenc}

\usepackage{microtype}

\usepackage{inconsolata}

\usepackage[ruled]{algorithm2e}
\usepackage{setspace}
\SetKwComment{Comment}{/* }{ */}

\usepackage{amsfonts}
\usepackage{amssymb}
\usepackage{pifont}
\newcommand{\cmark}{\ding{51}}%
\newcommand{\xmark}{\ding{55}}%
\usepackage{amsmath}
\usepackage{hyperref}
\usepackage[nameinlink]{cleveref}

\usepackage{booktabs}

\usepackage{graphicx}
\usepackage{float} 
\usepackage[subrefformat=parens]{subcaption}

\title{Jury: A Comprehensive Evaluation Toolkit}

\newcommand{\footremember}[2]{%
    \footnote{#2}
    \newcounter{#1}
    \setcounter{#1}{\value{footnote}}%
}
\newcommand{\footrecall}[1]{%
    \footnotemark[\value{#1}]%
} 

\author{
    Devrim Çavuşoğlu \footremember{obss}{OBSS AI} \footremember{metu}{Middle East Technical University} %
    \and Seçil Şen \footrecall{obss} \footremember{boun}{Bogazici University}  %
    \and Ulaş Sert \footrecall{obss} %
    \and Sinan Onur Altınuç \thanks{\hspace{0.1cm} Study is done while working at OBSS AI.} \footrecall{metu} \\%
    \footrecall{obss} \; OBSS AI \\
    \footrecall{metu} \; Middle East Technical University \\
    \footrecall{boun} \; Bogazici University \\
    \texttt{<first-name>.<last-name>@obss.tech}
}

\begin{document}
\maketitle
\begin{abstract}
Evaluation plays a critical role in deep learning as a fundamental block of any prediction-based system. However, the vast number of Natural Language Processing (NLP) tasks and the development of various metrics have led to challenges in evaluating different systems with different metrics. To address these challenges, we introduce \emph{jury}, a toolkit that provides a unified evaluation framework with standardized structures for performing evaluation across different tasks and metrics. The objective of \emph{jury} is to standardize and improve metric evaluation for all systems and aid the community in overcoming the challenges in evaluation. Since its open-source release, \emph{jury} has reached a wide audience and is available on GitHub at \href{https://github.com/obss/jury}{https://github.com/obss/jury}.
\end{abstract}

\section{Introduction}

Metrics play a crucial role, particularly in deep learning-based approaches. They are not only used to evaluate a trained model but also to gauge the progress of training and assess the quality of generalization. The evaluation in Natural Language Generation (NLG) is more challenging compared to other deep learning tasks due to the fact that assigning a numerical score to a textual output is a non-trivial problem. Thus, a research problem on this topic gains attention from most NLG researchers.

Earlier machine learning-based NLG research relied mostly on human evaluation since it is found to be the most straightforward way to measure \textit{human-likeliness} \cite{VANDERLEE2021101151}. The recent advances in deep learning push the state of the art in NLG research and the latest models come to a point where human evaluation is time and cost-intensive. Experimenting in deep learning to perfect models, datasets in terms of performance and quality mostly cost even more than training a single experiment in terms of time and resources. Most NLP research mainly relies on automatic evaluation metrics such as BLEU \cite{papineni-etal-2002-bleu}, METEOR \cite{banerjee-lavie-2005-meteor} and ROUGE \cite{lin-2004-rouge}.  Besides being fast, those metrics measure the linguistic and semantic quality of the generated textual output such as grammatical correctness, fluency, and diversity \cite{metricsurvey}. The use of automatic metrics for the assessment of a system’s quality, on the other hand, is controversial and has been criticized for not correlating well with human evaluation \cite{VANDERLEE2021101151}. Most studies combine the strengths of two evaluation techniques; human evaluation outputs are assumed to be the gold standard whereas the automatic evaluation metric results are
used to ensure fast iteration. 

NLP tasks possess inherent complexity, requiring a comprehensive evaluation of model performance beyond a single metric comparison. Established benchmarks such as WMT \cite{barrault-etal-2020-findings} and GLUE \cite{wang-etal-2018-glue} rely on multiple metrics to evaluate models on standardized datasets. This practice promotes fair comparisons across different models and pushes advancements in the field. Embracing multiple metric evaluations provides valuable insights into a model's generalization capabilities. By considering diverse metrics, such as accuracy, F1 score, BLEU, and ROUGE, researchers gain a holistic understanding of a model's response to never-seen inputs and its ability to generalize effectively. Furthermore, task-specific NLP metrics enable the assessment of additional dimensions, such as readability, fluency, and coherence. The comprehensive evaluation facilitated by multiple metric analysis allows for trade-off studies and aids in assessing generalization for task-independent models. Given these numerous advantages, NLP specialists lean towards employing multiple metric evaluations. 

Although employing multiple metric evaluation is common, there is a challenge in practical use because widely-used metric libraries lack support for combined and/or concurrent metric computations. Consequently, researchers face the burden of evaluating their models per metric, a process exacerbated by the scale and complexity of recent models and limited hardware capabilities. This bottleneck impedes the efficient assessment of NLP models and highlights the need for enhanced tooling in the metric computation for convenient evaluation. In order for concurrency to be beneficial at a maximum level, the system may require hardware accordingly. Having said that, the availability of the hardware comes into question. 

The extent of achievable concurrency in NLP research has traditionally relied upon the availability of hardware resources accessible to researchers. However, significant advancements have occurred in recent years, resulting in a notable reduction in the cost of high-end hardware, including multi-core CPUs and GPUs. This progress has transformed high-performance computing resources, which were once prohibitively expensive and predominantly confined to specific institutions or research labs, into more accessible and affordable assets. For instance, in BERT \cite{devlin-etal-2019-bert} and XLNet \cite{zhilin-etal-2019-xlnet}, it is stated that they leveraged the training process by using powerful yet cost-effective hardware resources. Those advancements show that the previously constraining factor for hardware accessibility has been mitigated, allowing researchers to overcome the limitations associated with achieving concurrent processing capabilities in NLP research.

To ease the use of automatic metrics in NLG research, several hands-on libraries have been developed such as \emph{nlg-eval} \cite{sharma2017nlgeval} and \emph{datasets/metrics} \cite{lhoest-etal-2021-datasets} (now as \emph{evaluate}). Although those libraries cover widely-used NLG metrics, they don't allow using multiple metrics in one go (i.e. combined evaluation), or they provide a crude way of doing so if they do. Those libraries restrict their users to compute each metric sequentially if users want to evaluate their models with multiple metrics which is time-consuming. Aside from this, there are a few problems in the libraries that support combined evaluation such as individual metric construction and passing compute time arguments (e.g. n-gram for BLEU \cite{papineni-etal-2002-bleu}), etc. Our system provides an effective computation framework and overcomes the aforementioned challenges.

We designed a system that enables the creation of user-defined metrics with a unified structure and the usage of multiple metrics in the evaluation process. Our library also exploits \emph{datasets} package to promote the open-source contribution; when users implement metrics, the implementation can be contributed to the \emph{datasets} package. Any new metric released by \emph{datasets} package will be readily available in our library as well. 

We introduce a novel comprehensive evaluation system, \emph{jury}, built on top of \emph{evaluate}, and our claim is that our strength comes from our three main capabilities that differ from existing toolkits. Our contribution can be summarized as follows: 

\begin{itemize}
    \item Unified interface for metric computations.  
    \item Combined metric evaluation.
    \item Evaluation of multiple predictions and/or multiple references. To the best of our knowledge, there is currently no software that supports this feature.
    \item Task mapping in metrics.
\end{itemize}  

Along with these contributions we build upon a powerful framework that reduces additional work for evaluation. Furthermore, it lets the community focus on their experiments, where the evaluation part is easily handled by \emph{jury}.

\section{Related Work}
In recent years, there has been a growing trend toward using automated evaluation methods for the assessment of NLP tasks. Several toolkits and libraries have been developed to provide easy-to-use and comprehensive solutions for evaluating NLP models.

\textbf{\emph{TorchMetrics}} \cite{Detlefsen2022} is a general-purpose metrics package that covers a wide variety of tasks and domains used in the machine learning community. The package provides standard classification and regression metrics, as well as domain-specific metrics for audio, computer vision, NLP, and information retrieval.

\textbf{\emph{nlg-eval}} \cite{sharma2017nlgeval} evaluation library that focuses on unsupervised automated metrics for NLG. This toolkit provides several metrics that can be used to evaluate the quality of NLG outputs, such as the BLEU score and the ROUGE score.

\textbf{\emph{datasets/metrics - evaluate}} \cite{lhoest-etal-2021-datasets} is another library that provides a more standardized and accessible method for evaluating and comparing models, as well as reporting their performance. The library offers dozens of popular metrics for a variety of tasks, including NLP and computer vision.



\section{Jury/Library Overview}

\begin{figure*}
\begin{subfigure}[b]{.44\linewidth}
\centering
\includegraphics[width=\linewidth]{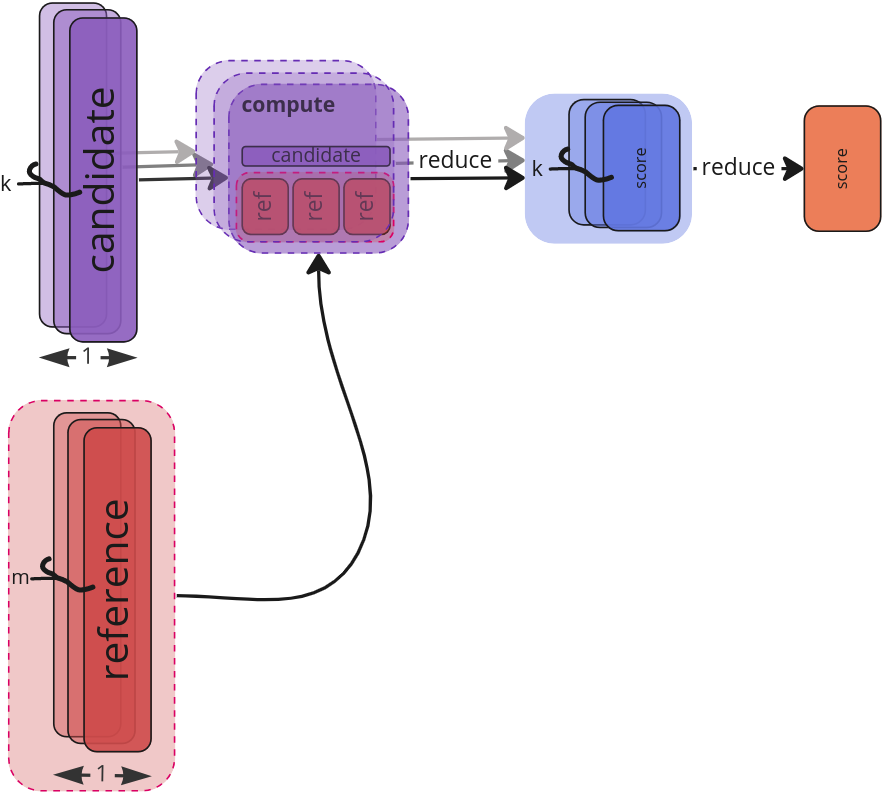}
\caption{Multiple predictions vs. multiple references for a single \emph{evaluation instance}.}\label{sfig:compute-a}
\end{subfigure}\hfill
\begin{subfigure}[b]{.44\linewidth}
\centering
\includegraphics[width=\linewidth]{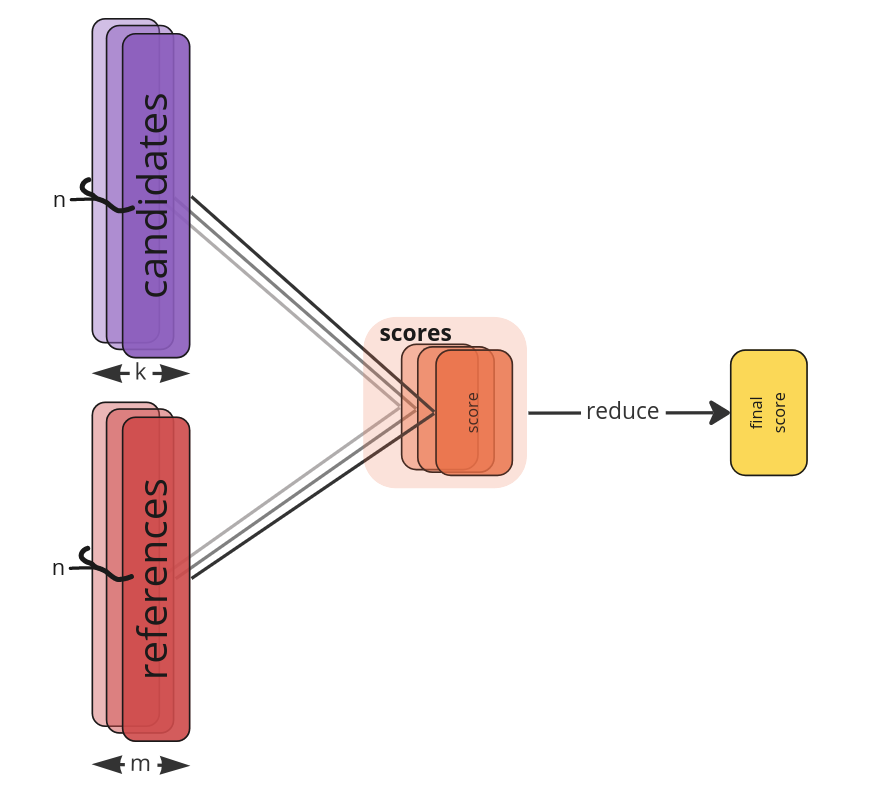}
\caption{Computation of metrics for multiple \emph{evaluation instances}.}\label{sfig:compute-b}
\end{subfigure}
\caption{General computation scheme of \emph{jury}. \textbf{(a)} Generally (and by default) both reduce operations are \emph{max}, optionally these can be determined separately. \textbf{(b)} The reduce is generally \emph{arithmetic-mean} for most of the metrics.}
\label{fig:compute}
\end{figure*}

As a starting point, instead of completely constructing some fundamental blocks from scratch, we chose \emph{datasets/metrics} \cite{lhoest-etal-2021-datasets} as a building block and built \emph{jury} on top of it. The toolkit \emph{datasets/metrics} are first released under the library \emph{datasets}, and later in 2022 \emph{datasets/metrics} is deprecated\footnote{\href{https://github.com/huggingface/datasets/pull/4739}{https://github.com/huggingface/datasets/pull/4739}} in favor of \emph{evaluate} which is an isolated library transferred from \emph{datasets/metrics}. We refactored our codebase to make a transition from \emph{datasets/metrics} to \emph{evaluate} accordingly. There are a few critical reasons that guided us to choose \emph{evaluate} to form the base structure and the skeleton of our system:

\begin{itemize}
    \item Having a base \verb|Metric| class that performs computations internally in a generic way which supports actions like appending, accumulating compute items over time.
    \item Utilizing arrow tables\footnote{Initially intended for storage and efficient computations on datasets, but also utilized for metrics). The arrow tables are useful also for verifying data types according to a schema for a table and forcing users to feed the correct input data type in calculations.} and efficient compression techniques.
    \item Huge community adoption and support.
    \item Easily add metrics to a community hub\footnote{Share metrics through HF spaces, so that metrics to be added do not necessarily have to be included in the library, and reduces the friction for custom metrics. \href{https://huggingface.co/metrics}{https://huggingface.co/metrics}} either public or private.
\end{itemize}

In light of these useful points, we formed our basis and internal architecture with \emph{evaluate}. Starting from this skeleton, we added functionalities and structures to further improve and build a more generic tool with the purpose of standardizing evaluation frameworks.

\subsection{Design \& Structures}
The library is developed with two core concepts in mind; a unified interface and generic use of metrics on different tasks.

The implemented metrics are facilitated with two main components \verb|Metric| and \verb|Jury| (high-level interface for all metric computations). \verb|Metric| is a base class that provides utility for tasks such as compute, collate (include pre-process), post-process, reduce, etc. to all metrics in all tasks while \verb|Jury| provides computing multiple metrics for the same input concurrently and creates a unified report after the computations.

\subsection{Metric}
We adopted \verb|evaluate.Metric| as a base class for our metric implementations to inherit the useful properties (e.g. efficient computations, caching, utilization of vectorization, etc.) and extended this to form our \verb|Metric| class.

General computation scheme of metric computations is given in \autoref{fig:compute}.

\subsubsection{Unified Interface}
In NLG tasks we may want to evaluate the systems with multiple references per instance if resources are available. The most prominent practice of this is observed in machine translation tasks, where given input in a source language the system tries to give a corresponding output in a target language. While evaluating such a system, it is required that a source sentence instance is paired with a reference target sentence. In fact, a single source sentence could be paired with different target sentences that convey the same meaning and reflect the true structure of the source language in the target language as much as possible. The multiple references are useful as a single reference might not convey the true variability of true language variations, and there have been studies on the effect of multiple references \cite{thompson1991automatic, turian-etal-2003-evaluation, qin-specia-2015-truly-exploring}. For this reason, there are metric implementations and proposals that involves multiple references in a correspondence to a single prediction (a.k.a candidate or hypothesis) \cite{post-2018-call, zhang2020bertscore}.

The practice of supporting multiple references, however, is not adopted in the implementations of all metrics in available libraries and frameworks. For example, in \emph{evaluate} only some subset of metrics accept multiple references. We have introduced a unified interface for all metrics implemented under \emph{jury} to support multiple references. For each metric, the \verb|compute| method now takes a set of reference collections, allowing users to pass either a single collection or multiple collections of varying lengths.

The importance of multiple predictions in evaluation phases has been recently studied \cite{fomicheva-etal-2020-multi, liu-etal-2021-neural}. The study of multi-hypothesis MT evaluation \cite{fomicheva-etal-2020-multi} suggests that exploiting multi-hypothesis has achieved a higher correlation with human judgement. There are several methods for obtaining multiple predictions from a single input, such as perturbations, sampling techniques, and post-processing. These methods can create variations in model outputs that can be used for evaluation, quality assessment (e.g. for ranking predictions and refining system outputs), and other purposes. Hence, along with adding support for multiple references for all metrics implemented in \emph{jury}, we also included the usage of multiple predictions in the metrics in the library, and established a unified metric computation interface for all metrics.

The computation for metrics is carried out in a generic framework that is illustrated in \autoref{fig:compute}. Basically, \emph{jury} computes prediction instances against paired reference instances and uses reduce operations to obtain a single score for each evaluation instance. Then, it generally computes the arithmetic mean of those scores to obtain the final score at the evaluation collection/corpus level. The computation framework is detailed in \autoref{apx:computation}.

\subsubsection{Task Assigned Metrics}
The system's outputs naturally vary among different tasks. The most apparent example to observe this difference is between Natural Language Generation (NLG) and Natural Language Inference (NLI). In NLG systems, such as machine translation, abstractive text summarization, and question generation, the final output is typically a string or a set of strings. However, in NLI systems the final output is an integer that corresponds to a label id. Apart from that, while NLI is inherently a single-label multi-class classification task, a generic text classification could very well be a multi-label multi-class classification task \cite{lewis2004rcv1}. These differences between tasks dictate the type of outputs. Therefore, to handle different system outputs based on different tasks, we introduced a way of mapping the metric to the computation of that particular task (e.g. \emph{accuracy-for-language-generation}, \emph{accuracy-for-sequence-classification}).

The default task is determined as language generation (i.e. NLG) for all metrics where such metrics accept string data type. Though most of the metrics accept text string as input, some metrics (e.g. BLEU) accept tokenized input. To standardize the mapping of this input type, we included an optional tokenizer argument for BLEU which can also be included for other potential metrics requiring tokenized input.

With this task mapping management, the computations are less error-prone and much easier without user requiring to build a custom pre or post processing. Also, a new task mapping can be added easily to the library, or defined as a custom class for personal use by extending \verb|MetricForTask|. Currently \emph{jury} has four main task classes; \verb|MetricForLanguageGeneration|, \verb|MetricForSequenceClassification|, \verb|MetricForSequenceLabeling| and \verb|MetricForCrossLingualEvaluation|. These metric classes can be extended and custom metrics can be created easily. Additionally, the \verb|MetricForTask| class can be extended to support other tasks.

\subsection{Scorer}
 We designed a high-level interface with a class named \verb|Jury| which allows computing many metrics at once, either concurrently or sequentially. When the computations are over, it outputs a well-formed dictionary where each score is reported with additional information. 

In addition, this component enhances the reliability of the evaluation by verifying that the metrics being computed are appropriate for the task at hand. It only allows metrics that belong to the same task used together, thereby reducing the potential risks (e.g. miscalculation, misinterpretation) are avoided.

\begin{figure*}
\begin{subfigure}[t]{.45\linewidth}
\centering
\includegraphics[width=\linewidth]{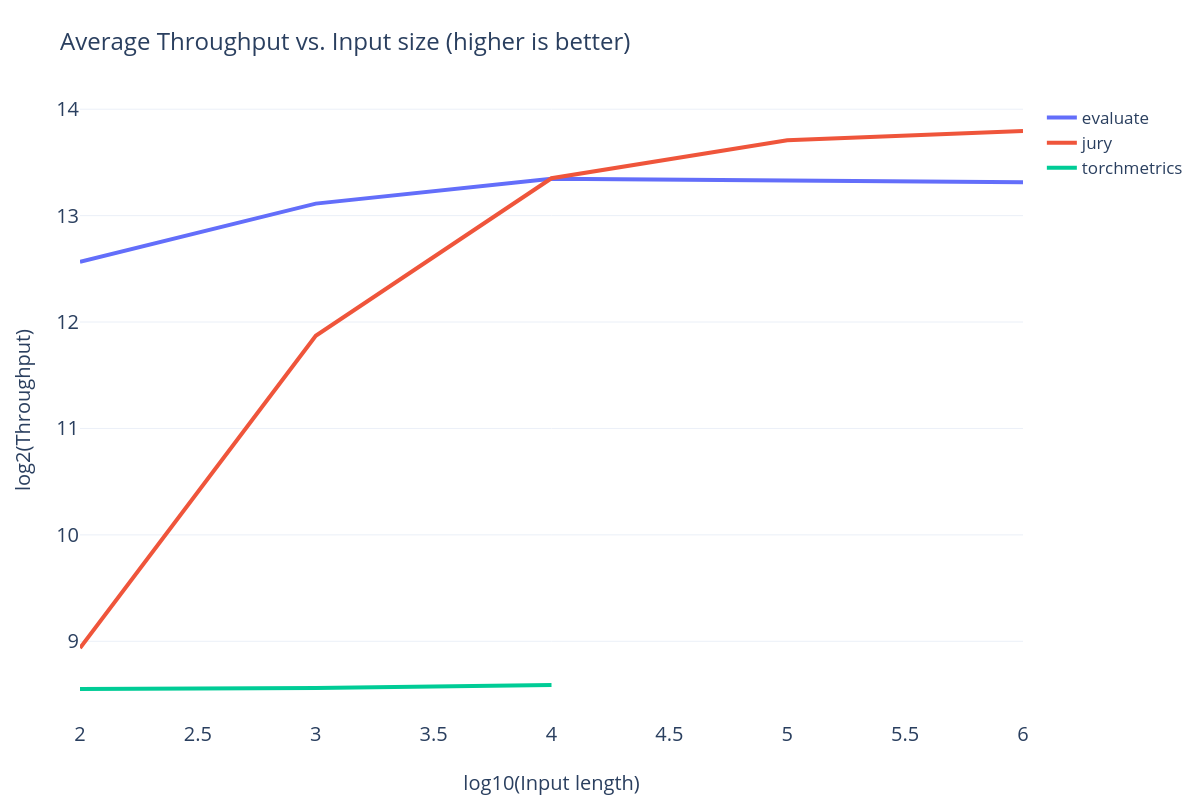}
\caption{Different evaluation tools scale against input size (number of instances). The x-axis is $log_{10}$ scaled.}\label{sfig:inputsize}
\end{subfigure}\hfill
\begin{subfigure}[t]{.45\linewidth}
\centering
\includegraphics[width=\linewidth]{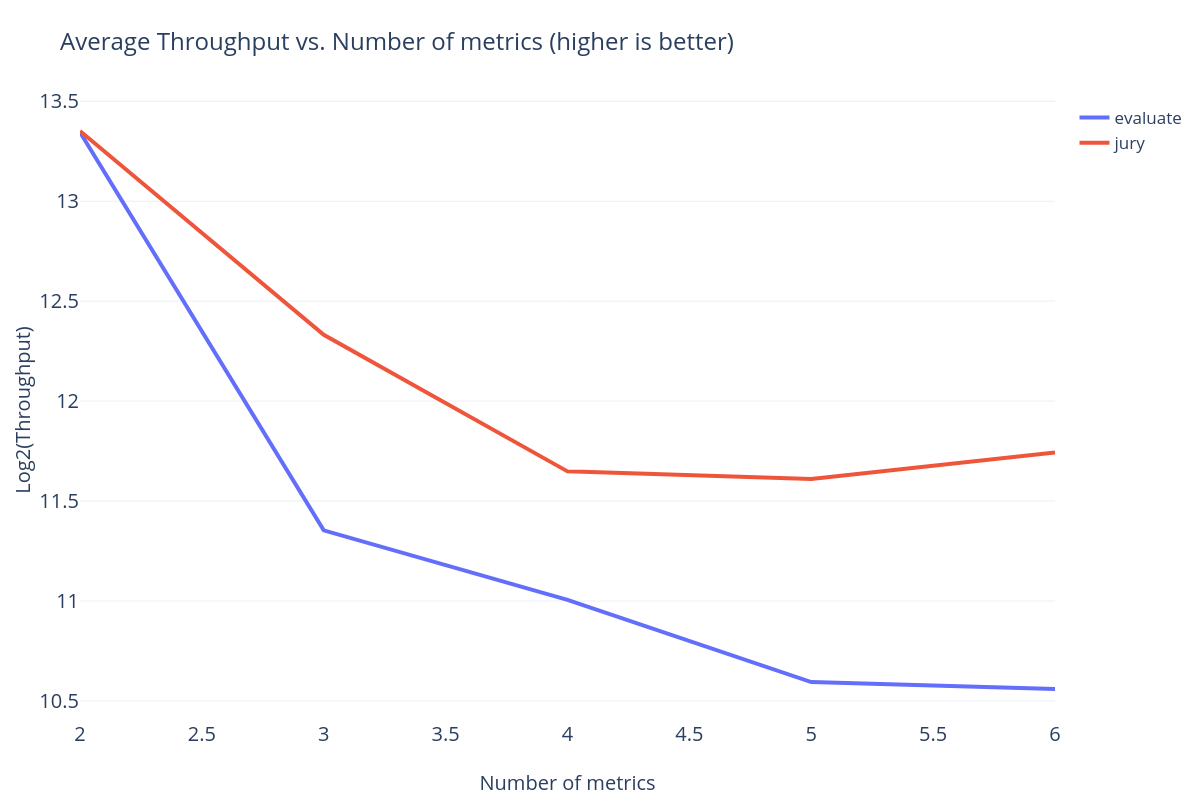}
\caption{Different evaluation tools scale against number of metrics.}\label{sfig:nummetrics}
\end{subfigure}
\caption{Experiments conducted against other open-source frameworks where \textbf{(a)} compares how well the tools scale in terms of input size and \textbf{(b)} compares how well the tools scale in terms of number of metrics. The evaluation is done on throughput (items/s) which is scaled by $log_2$ in both of the plots. The results shown in figures are average of 5 runs.}
\label{fig:comparison}
\end{figure*}

\section{Library Tour}

Loading a single metric is a simple one-liner adopted from \emph{evaluate} where the interface is the same except few additional keyword arguments.

\begin{figure}[H]
\centering
\includegraphics[width=\linewidth]{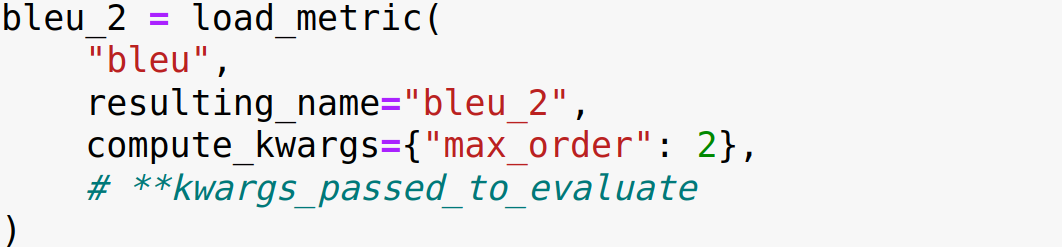}
\label{ilfig:load_metric}
\end{figure}
\vspace{-0.6cm}

Computation for a single metric is carried out by the \verb|compute| method of a metric object.

\begin{figure}[H]
\centering
\includegraphics[width=\linewidth]{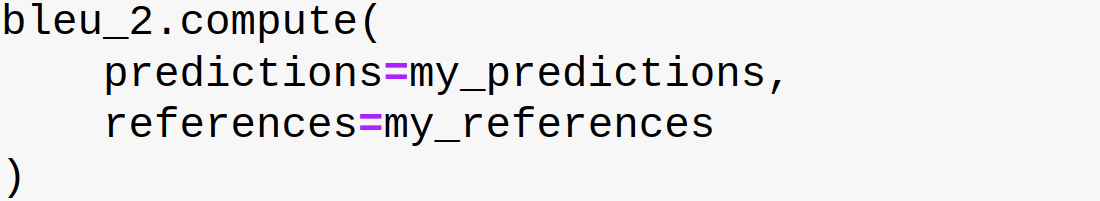}
\label{ilfig:metric_compute}
\end{figure}
\vspace{-0.6cm}

Obtaining scores through \verb|Jury| class (scorer) allows combining multiple metrics and allows concurrent computation.

\begin{figure}[H]
\centering
\includegraphics[width=\linewidth]{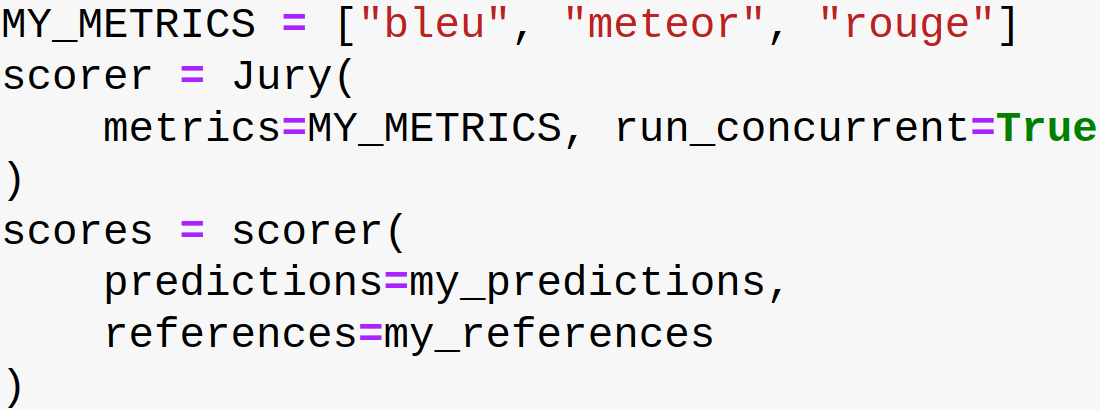}
\label{ilfig:jury_compute}
\end{figure}
\vspace{-0.6cm}

The usage examples provided above involve loading a single metric and computing via the \verb|compute()| method, or using the \verb|Jury| class and calling it with predictions and references. To minimize the risk of errors, users must pass keyword arguments with explicit names when using either of these methods.

\section{Discussion}

\begin{table*}
\centering
    \begin{tabular}{l|ccccc}
        \toprule
         & Jury & Datasets & Evaluate & NLGEval & Torchmetrics\\
        \midrule
        Metrics & $19 \cup Datasets $ & 22 & 22 & 8 & 13\\
        Combined evaluation & \cmark & \xmark & \cmark & \xmark & \xmark\\
        Concurrency & \cmark & \xmark & \xmark & \xmark & \xmark\\
        Task mapping & \cmark & \xmark & \xmark & \xmark & \xmark\\
        Unified I/O & \cmark & \xmark & \xmark & \xmark & \xmark\\
        \bottomrule
    \end{tabular}
\caption{Feature matrix comparing publicly available evaluation packages.}
\label{tab:feature}
\end{table*}

We introduce and improved the base \emph{evaluate} library by standardizing the interface with our unified interface design and task mapping for metrics. Additionally, we added support for combining multiple metrics to compute concurrently, which reduces the run time depending on the hardware for users. We conducted experiments on runtime of the computations for evaluations by taking two common open-source evaluation libraries; namely \emph{torchmetrics} and \emph{evaluate}. The results of the experiments are given in \autoref{fig:comparison}. The details corresponding to the experiments are given in \autoref{apx:comparison}. Also, feature-wise comparison is given in \autoref{tab:feature}. 

We also added a sanity-checking mechanism that checks the viability of the metrics passed to \verb|Jury| scorer whether the given metrics are appropriate to the same task or not. In addition to these, \emph{jury} seamlessly supports all metrics from \emph{evaluate} as is as instructed in \emph{evaluate}.

Unfortunately, however, if a metric is not implemented in \emph{jury} with a unified structure, it cannot be used with the same interface as the \emph{jury} metrics (e.g., passing multiple predictions). This is the most noticeable limitation of \emph{jury} as of the time being. To encounter this problem, we put a good amount of effort to make the custom metric implementation easy as possible, so that it can be easily alleviated by users.

Lastly, the underlying type checking on arrow tables enforces a rigid expectation on metric input types and formats. This results in reduced flexibility for modifying metric inputs and a slightly increased learning curve for new users. However, in exchange, this feature contributes to code safety by ensuring error detection, which in turn improves the overall quality of the code. Additionally, the performance is also enhanced, with the type checking to facilitate more efficient processing and execution of the code. Thus, arrow tables' trade-offs are outweighed by their contribution to code safety, quality, and performance.

\section{Conclusion}
We introduced \textit{jury} a comprehensive evaluation toolkit for NLP metrics, main focus in development is on being extensible and customizable. Evaluation in NLP, particularly in natural language generation tasks, is a challenging task due to a number of practical difficulties that can arise. These difficulties include but not limited to \textbf{(i)} long run-time in computation of multiple metrics sequentially, \textbf{(ii)} there is no standard in input type of metric computation (some take single prediction instance, some allows multiple references, etc.), \textbf{(iii)} there is no standard of task mapping in metrics (i.e. a metric can be used for different tasks with a different computation scheme). While \emph{jury} addresses all of these issues, it also extends the usefulness of the library in the following ways:
\begin{itemize}
    \item \textbf{(easy-to-use)} Providing easy access to already available and well-maintained metrics through a unified interface
    \item \textbf{(generic input type)} Providing support for predictions with multiple predictions and/or multiple references
    \item \textbf{(task mapping)} Utilizing metrics among different tasks
    \item \textbf{(accelerated run-time)} Providing faster and more efficient evaluations with concurrent metrics calculation and taking advantage of the arrow table structure
\end{itemize}

Our primary objective is to simplify, accelerate, and enhance the evaluation of NLP tasks by providing researchers with convenient access to a broad range of evaluation metrics. By automating the evaluation process, researchers can analyze and improve their models more efficiently, saving valuable time and resources.

\section*{Acknowledgements}
We would also like to express our appreciation to Cemil Cengiz for fruitful discussions.

\bibliography{main}
\bibliographystyle{acl_natbib}

\appendix

\section{Computation Schema of Metrics}
\label{apx:computation}

\begin{algorithm*}
\setstretch{0.7}
\SetAlgoLined
\caption{Generic computation framework for a single metric. $[ ]$ represents an empty array.}
\label{alg:compute}
\KwData{$C_{eval}$}
\KwResult{$metric \; score$}
$N \gets |C_{eval}|$\;
$S_{all} \gets [ \; ]$\;
\For{$n \gets 1$ to $N$}{
    $(I_{pred}^{(n)}, I_{ref}^{(n)}) \gets C_{eval}^{(n)}$\;
    $K \gets |I_{pred}^{(n)}|$\;
    $M \gets |I_{ref}^{(n)}|$\;
    $S_{instance} \gets [ \; ]$\;
    \For{$k \gets 1$ to $K$}{
        $p \gets I_{pred,k}^{(n)}$\;
        $S_{pred} \gets [ \; ]$\;
        \For{$m \gets 1$ to $M$}{
            $r \gets I_{ref,m}^{(n)}$\;
            $S \gets metric(p, r)$\;
            \text{add $S$ to $S_{pred}$}\;
        }
        $S' \gets reduce^{(1)}(S_{pred})$\;
        \text{add $S'$ to $S_{instance}$}\;
    }
    $S'' \gets reduce^{(2)}(S_{instance})$\;
    \text{add $S''$ to $S_{all}$}\;
}
$S_{final} \gets reduce^{(3)}(S_{all})$ \tcp*{$reduce^{(3)}$ is generally an arithmetic mean}
\Return{$S_{final}$}\;
\end{algorithm*}

For convenience, we will call a single prediction-reference instance an \emph{evaluation instance}, $I_{eval}$, which is composed of a paired collection of predictions obtained from the same input and a collection of references that corresponds to the same source for a single data point. Multiple references in a single instance collection will be referred to as a single \emph{reference instance}, $I_{ref}$, and likewise, multiple predictions in a single instance collection will be referred to as a single \emph{prediction instance}, $I_{pred}$, regardless of the length of the collection. The collection of evaluations instances which serves as the input to the metric computation is referred as \emph{evaluation collection}, $C_{eval}$. The input for predictions is a set of \emph{prediction instances}, $C_{pred}$, and references is a set of \emph{reference instances}, $C_{ref}$, (e.g. sequence of sequence of string). The schema for the instances is given in \autoref{eq:instances}.

\begin{equation}
\label{eq:instances}
    \begin{gathered}
        I_{pred}^{(i)} = \{ p_1^{(i)}, p_2^{(i)}, \dots, p_k^{(i)} \} \; k \in \mathbb{N}\\
        I_{ref}^{(i)} = \{ r_1^{(i)}, r_2^{(i)}, \dots, r_m^{(i)} \}  \; m \in \mathbb{N}\\ 
        I_{eval}^{(i)} = (I_{pred}^{(i)}, I_{ref}^{(i)}) \; \forall \, i=0,1,2,\dots,n \\
        C_{pred} = \{ I_{pred}^{(1)}, I_{pred}^{(2)}, \dots, I_{pred}^{(n)} \} \\
        C_{ref} = \{ I_{ref}^{(1)}, I_{ref}^{(2)}, \dots, I_{ref}^{(n)} \} \\
        C_{eval} = \{ I_{eval}^{(1)}, I_{eval}^{(2)}, \dots, I_{eval}^{(n)} \} \; n \in \mathbb{N}
    \end{gathered}
\end{equation}

 The generic framework for metric computations incorporating multiple predictions and multiple references is given in \Cref{alg:compute}. The first phase of the computation is computing the metric for a single evaluation instance. For each prediction in a \emph{prediction instance} (of length $k$) is compared against corresponding \emph{reference instance} (of length $m$), where the length of \emph{prediction instance} and \emph{reference instance} do not have to be the same length. In the first phase of computation, the metric score for each prediction in the \emph{prediction instance} is computed by comparing it against all references in the corresponding \emph{reference instance}, resulting in a total of $m$ scores, i.e length of \emph{reference instance}. These scores are then reduced to a single score where the default option is \emph{max}. This operation is carried out for each prediction in the \emph{prediction instance}, so in total, we get scores of length $k$, and we reduce this to a single score (default option is \emph{max}) to obtain a score for that \emph{evaluation instance}. This scheme is illustrated in \autoref{sfig:compute-a}.

Assume that we have $n$ \emph{evaluation instances} in total, we apply the computations described above and get a single score per instance, and thus $n$ scores in total, then we reduce it again to obtain the overall score. Although most of the metrics are reduced by \emph{arithmetic-mean}, some metrics (e.g. BLEU) requires additional handle to fit the framework to multiple instances. For example, BLEU takes translation length and reference length into account while computing the score. The computation scheme for the second phase is given in \autoref{sfig:compute-b}.

\section{Experiments on Runtime}
\label{apx:comparison}

The experiments regarding the comparison of runtimes of the tools, i.e. \emph{torchmetrics v0.11.4} and \emph{evaluate v0.4.0}, against \emph{jury v2.2.4} are conducted with the following setup:

\begin{itemize}
    \item The experiments are conducted in a machine having a Linux operating system (Ubuntu 22.04.1 LTS) and with 12 core processor "Intel(R) Core(TM) i7-10750H CPU @ 2.60GHz".
    \item Experiments are conducted with Python 3.9.16.
    \item The results reported in the plots are average of 5 runs. In both of the plots in \autoref{fig:comparison} the throughput (number of evaluation instances per second) is used for comparison, and it is $log_2$ scaled for better interpretibility.
\end{itemize}

\textit{\textbf{Experiment I}} For the results in \autoref{sfig:inputsize}, scorer with 2 metrics are used for all packages, the metrics are selected as \textit{BLEU} \cite{papineni-etal-2002-bleu} and \textit{SacreBLEU} \cite{post-2018-call}. The number of metrics for scorer is fixed at 2 for different input sizes. Since the difference for \emph{torchmetrics} vs both \textit{evaluate} and \textit{jury} are significantly inferior, no further experiments are conducted after $10^4$ input size for \emph{torchmetrics}.

\textit{\textbf{Experiment II}} For the results in \autoref{sfig:nummetrics}, scorer has updated by adding a single metric each time from 2 to 6 metrics. The metric setup starting from 2 metrics as in \textit{Experiment I}, and \textit{METEOR} \cite{banerjee-lavie-2005-meteor}, \textit{TER} \cite{snover-etal-2006-study}, \textit{CHRF} \cite{popovic-2015-chrf}, \textit{GoogleBLEU} \cite{wu2016google} are added in order step-by-step. The input size is fixed at $10,000$ evaluation instances.  From the insight of \textit{Experiment I}, \emph{torchmetrics} is not included in this setup as its performance is not comparable in speed to jury and evaluate.

\end{document}